\documentclass[10pt,twocolumn,letterpaper]{article}

\usepackage[pagenumbers]{cvpr} 

\definecolor{cvprblue}{rgb}{0.21,0.49,0.74}
\usepackage[pagebackref,breaklinks,colorlinks,allcolors=cvprblue]{hyperref}
\usepackage[T1]{fontenc}
\usepackage{multirow}
\usepackage{algorithm}
\usepackage{algorithmic}

\def\paperID{1856} 
\def\confName{CVPR}
\def\confYear{2025}

\title{Facial Features Matter: a Dynamic Watermark based Proactive Deepfake Detection Approach}

\author{
Shulin Lan$^1$, Kanlin Liu$^2$, Yazhou Zhao$^2$, Chen Yang$^2$,\\ Yingchao Wang$^2$, Xingshan Yao$^2$, Liehuang Zhu$^2$\\
$^1$University of Chinese Academy of Sciences $^2$Beijing Institute of Technology\\
\tt\small lanshulin@ucas.ac.cn, \{3220211116, 3220242012, yangchen666\}@bit.edu.cn \\
\tt\small \{yingchaowang, 3120245900, liehuangz\}@bit.edu.cn
}

\begin{document}
\maketitle
\begin{abstract}
    Current passive deepfake face-swapping detection methods encounter significance bottlenecks in model generalization capabilities. Meanwhile, proactive detection methods often use fixed watermarks which lack a close relationship with the content they protect and are vulnerable to security risks. Dynamic watermarks based on facial features offer a promising solution, as these features provide unique identifiers. Therefore, this paper proposes a \underline{F}\underline{a}\underline{c}ial F\underline{e}ature-based \underline{P}\underline{r}\underline{o}active deepfake de\underline{t}\underline{e}\underline{c}\underline{t}ion method (FaceProtect), which utilizes changes in facial characteristics during deepfake manipulation as a novel detection mechanism. We introduce a GAN-based One-way Dynamic Watermark Generating Mechanism (GODWGM) that uses 128-dimensional facial feature vectors as inputs. This method creates irreversible mappings from facial features to watermarks, enhancing protection against various reverse inference attacks. Additionally, we propose a Watermark-based Verification Strategy (WVS) that combines steganography with GODWGM, allowing simultaneous transmission of the benchmark watermark representing facial features within the image. Experimental results demonstrate that our proposed method maintains exceptional detection performance and exhibits high practicality on images altered by various deepfake techniques.
\end{abstract}

\begin{figure*}[ht]
    \centering
    \includegraphics[width=1.0\linewidth]{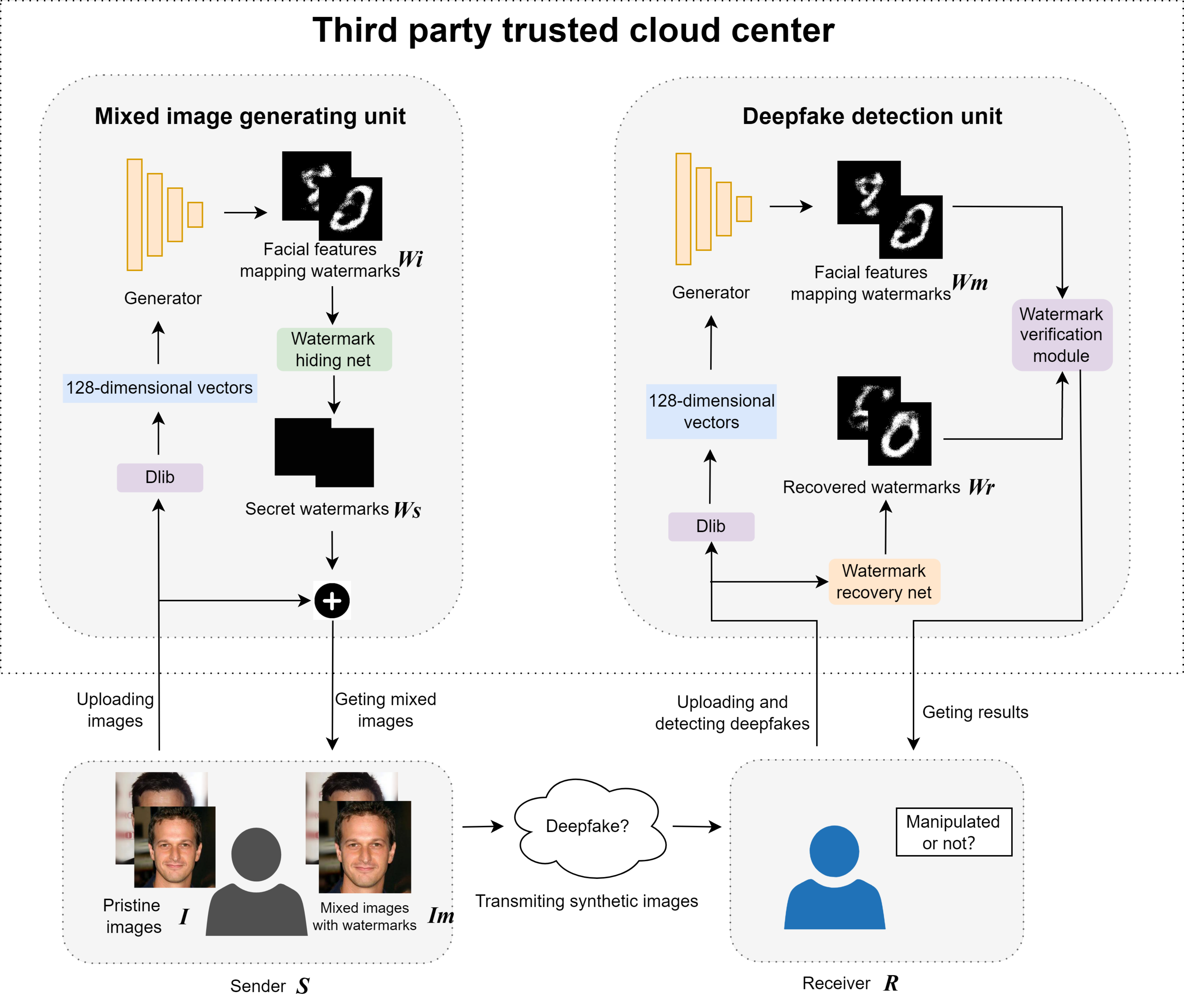}
    \caption{Overview of the proactive detection framework based on differences in facial features.}
    \label{fig:framework}
\end{figure*}
    
\section{Introduction}
\label{sec:intro}
The widespread dissemination of false information on social media through face-swapping has led to numerous adverse effects on everyday life. Moreover, deepfakes are frequently used to produce adult films and defraud acquaintances, endangering both individuals and communities. Consequently, it is essential to develop accurate and efficient methods for detecting deepfakes.

Passive detection techniques encounter significant challenges, particularly in their ability of generalization. They are seriously threatened by pseudo-face images that bear striking similarities to real faces in terms of texture and semantic characteristics, thanks to recent advancements in generative adversarial networks (GAN)\cite{GAN, wgan, stylegan2,starganv2, ganp, lgan, cyclegan}. To improve detection performance, current deepfake detection strategies concentrate on \textbf{broadening the variety of training data} \cite{facexray, sbi} or \textbf{enhancing related detection networks or algorithms}\cite{shallownetwork,capsulenetwork}. Nevertheless, these methods are susceptible to overfitting certain tamper traces or artifacts\cite{sbi}  (\eg, landmark mismatch, color discrepancies, and frequency inconsistencies). As a result, detection accuracy on cross-tampering method datasets is significantly reduced.

In light of this, researchers have incorporated \textbf{watermark technology} into deepfake detection. This \textbf{proactive detection} involves \textbf{embedding a watermark} as a representation of image interity. Some approaches\cite{agf,faketracer, agfv2} inject fingerprints or traces into training datasets, combining the data(fingerprints or traces) with parameters of specific image-generating networks, and recovering the data to detect alterations. Other methods\cite{antiforgery, cuma,visiblewatermark} prevent attackers from producing visually perceptible fake images by introducing noise or disruption. However, these aforementioned approaches are often limited to particular networks and scenarios, implying that retraining is required when certain components (\eg, the fingerprint, dataset, or network) change, which hampers pracical implementation. In addition, from a security perspective, current proactive detection approaches\cite{faceguard, facesign} primarily employ fixed sequences as watermarks, which is inadequate. While some mothods\cite{identitywatermark} introduce random sequences, they do not provide a viable technique for synchronizing random sequences between sender and recipient. Instead, they assume that both parties know the random sequence exactly. However, the significance of \textbf{facial features} as indicators of facial forgery has been overlooked in current research. These features can be used to create a distinctive representation of faces, even in twins with highly similar appearances. This observation presents an opportunity for developing more robust and generalizable deepfake detection methods.

In this paper, we address current challenges in deepfake detection by proposing a novel proactive detection framework (Fig.\ref{fig:framework}). The followings are the main contributions of this article.
\begin{itemize}
 \item[\textbullet] \textbf{\underline{F}\underline{a}\underline{c}ial F\underline{e}ature-based \underline{P}\underline{r}\underline{o}active deepfake de\underline{t}\underline{e}\underline{c}\underline{t}ion method (FaceProtect):}
 We utilize facial features as verification data, departing from conventional methods. This approach capitalizes on the unique characteristics of individual faces to enhance detection accuracy.
 \item[\textbullet] \textbf{GAN-based One-way Dynamic Watermark Generating Mechanism (GODWGM):}
 We design a GAN-based one-way dynamic watermark generating mechanism to transform facial features into associated watermarks. This innovative system employs 128-dimensional facial feature vectors as generator input instead of random noise , creating a one-way mapping from facial features to watermarks. This approach enhances security against reverse engineering. Moreover, it will not affect additional images as the watermark changes dynamically with the input image.
 \item[\textbullet] \textbf{Watermark-based Verification Strategy (WVS):}
 We introduce a watermark-based verification strategy(WVS) to solve watermark synchronization issues. It utilizes a trained hiding and recovery network based on U-Net\cite{unet} and SENet\cite{senet}, which can effectively hide watermarks in pristine images while maintaining their visual consistency. In addition, the recovery of watermarks from the mixed images(pristine images with invisible watermarks) is achieved using an advanced convolutional neural network.
 \item[\textbullet] \textbf{Advanced Deepfake Detection Method:}
 Our method introduces a novel approach to deepfake detection that operates independently of watermark synchronization. By computing the cosine similarity between the restored watermark and the mapped watermark, an approach not considered in previous work.
\end{itemize}

\begin{figure*}[ht]
    \centering
    \includegraphics[width=1.0\linewidth]{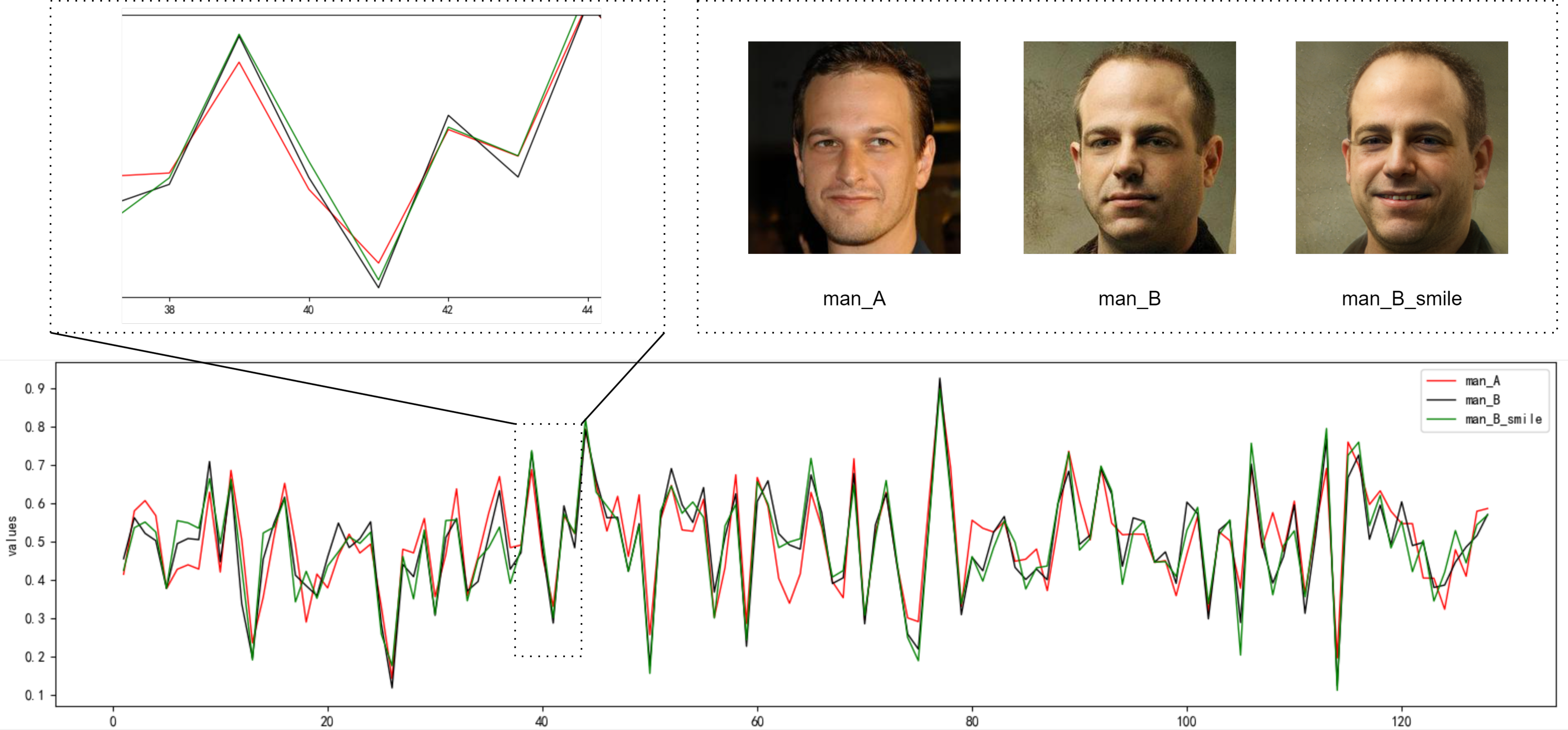}
    \caption{Curves of different people with similar facical features and the same person with different expressions.}
    \label{fig:intuition}
\end{figure*}
\section{Related Work}
\label{sec:relate}
\textbf{Passive Deepfake Detection.} A common approach in passive detection is to enrich images in training datasets with various forgery artifacts or optimize the network to improve detection performance. Several studies\cite{facexray,sbi,pcl,selfsupervised}show how to create training datasets with more forgery artifacts to provide the model with a more robust and general detection capability. In \cite{multiattention}, attention machanisms are used to guide the models in focusing on subtle distinctions between real and artificial faces. In \cite{rnn}, deepfake detection is achieved through the deployment of a recurrent convolutional network. \cite{Lipsnotelllies} learns lip reading and detects forgery artifacts by analyzing abnormal lip movements using a spatial-temporal network. Another approach is to capture the frequency information of images. $F^3$-Net\cite{thinkinginfrequency}, developed after analyzing the frequency characteristics of fake images, employs a learnable adaptive frequency feature extraction module. This module is also used in \cite{frequencyawareDF} to extract differential characteristics from various frequency bands. However, these methods primarily improve detection performance for compressed videos or images. Modal complementarity can be employed as a third strategy to enhance detection performance. According to\cite{vftelldeepfake}, voices and faces are proved to be significantly connected. \cite{dgm4} utilizes multimodal information to locate tampered content in addition to determining the authenticity of the input image text pair. However, these methods require text or audio in addition to images. 

\textbf{Proactive Deepfake Defense.} Before posting an image, multiple processes are carried out in order to accomplish proactive deepfake detection. These operations can be divided into two categories. The first is to add watermarks (primarily binary sequences) to images through steganography\cite{agf, faketracer, agfv2,faceguard,facesign,identitywatermark}. By recovering the sequence from the target image and comparing it with the original one, a specific similarity threshold is met, signifying that the image is genuine. Another approach involves embedding specific information, such as targeted noise or perturbations, into original images to prevent deepfakes from producing high-quality images\cite{antiforgery, cuma, 2step}. \cite{ visiblewatermark} makes the forged images display distinct labels, which facilitates detection. However, these methods have certain drawbacks. The use of fixed watermarks makes them more vulnerable to security risks, and there is no practical solution to synchronize benchmark watermarks for comparison. Additionally, adding specific information is only effective for particular generating models, and strong resistance is likely to develop if new deepfakes are encountered.

\section{Methodology}
\label{sec:methodology}
\begin{figure*}[ht]
  \centering
  \includegraphics[width=1.0\linewidth]{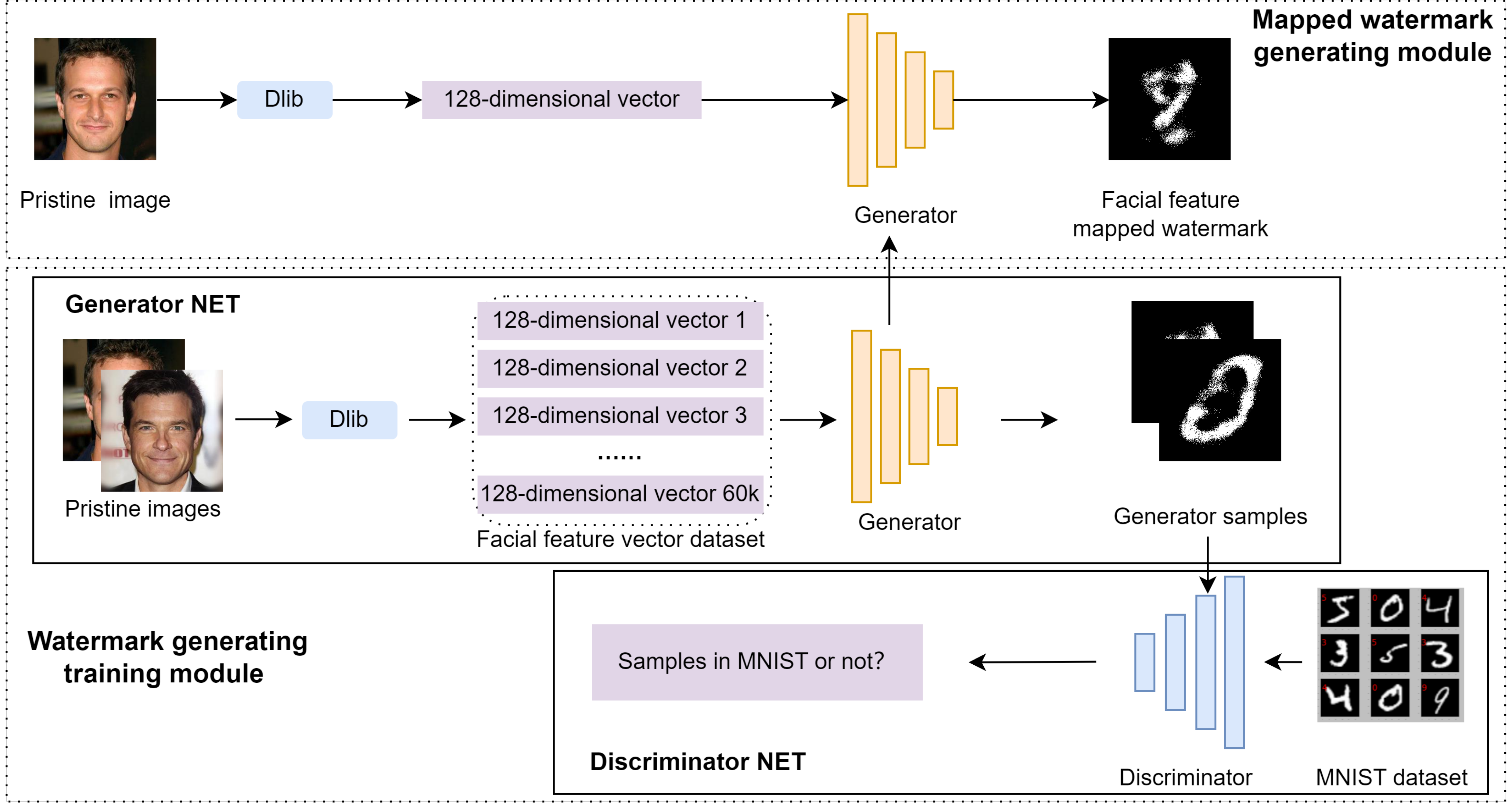}
  \caption{Overview of GAN-based one-way dynamic watermark generating mechanism(GODWGM).}
  \label{fig:godwgm}
\end{figure*}
\subsection{Method Intuition}
Deepfakes \eg, face swapping or face attribute modification, inherently alter facial features to achieve convincing results. Consequently, the distinction between a face image's pre- and post-tampering attributes can be used to identify deepfakes. We use Dlib\cite{dlib} to extract 128-dimensional facial feature vectors of different people with similar faces and the same person with different facial expressions. The curves of these facial features differ significantly, as shown in the bottom part of Fig.\ref{fig:intuition}. We focus on the regions where the curves nearly overlap, as illustrated in the upper left portion of Fig.\ref{fig:intuition}. Notably, even in areas with similar facial features, subtle distinctions between the curves are evident. These subtle variations in facial features provide a robust foundation for differentiating between manipulated and authentic faces.
\subsection{\texorpdfstring{%
    \underline{F}\underline{a}\underline{c}ial F\underline{e}ature-based \underline{P}\underline{r}\underline{o}active Deepfake De\underline{t}\underline{e}\underline{c}\underline{t}ion Method (FaceProtect)
}{%
    Facial Feature-based Proactive Deepfake Detection Method(FaceProtect)
}}
Based on the aforementioned findings, we present a proactive detection framework illustrated in Fig.\ref{fig:framework}. The proposed framework consists of three main components: the image owner, the receiver, and a trusted, authoritative third-party cloud center. The cloud center deploys two core modules: a mixed image generation unit and a deepfake detection unit. The mixed image generation unit is responsible for embedding watermark images associated with facial features into original images, producing mixed images that maintain visual consistency with the originals. The deepfake detection unit recovers the watermark image from the mixed image and compares it with the watermark image mapped from the facial features of the mixed image, thereby enabling deepfake detection. These two units are based on the GAN-based One-way Dynamic Watermark Generating Mechanism (GODWGM) and the Watermark-based Verification Strategy (WVS) proposed in this paper, which provide the technical and theoretical foundation for the framework. Initially, the original image $I$ is passed from the image holder $S$ to the mixed image generating unit. This unit incorporates a GAN-based one-way dynamic watermark generating mechanism. By mapping facial features to a grayscale image as the watermark $W_i$, it addresses security issues associated with typical fixed watermarks. The watermark dynamically adjusts in response to changes in the input image. Moreover, the generation process involves a one-way mapping. The secret watermark $W_s$ is generated by inputting the watermark $W_i$ into the steganography hiding network. $W_s$ is then combined with the pristine image $I$ to create a mixed image $I_m$, which is subsequently sent to the receiver $R$. Next, the receiver $R$ uploads $I_m$ to the deepfake detection unit, which employs a watermark-based verification strategy (WVS). It extracts the facial feature-mapped watermark $W_m$ from $I_m$ using GODWGM. Simultaneously, the hidden watermark $W_r$ of $I_m$ is obtained using the recovery network in the WVS. Both watermarks, $W_m$ and $W_r$ are then input into the watermark verification module for similarity comparison. Ultimately, the detection result is acquired. 
\begin{figure*}[ht]
  \centering
  \includegraphics[width=1.0\linewidth]{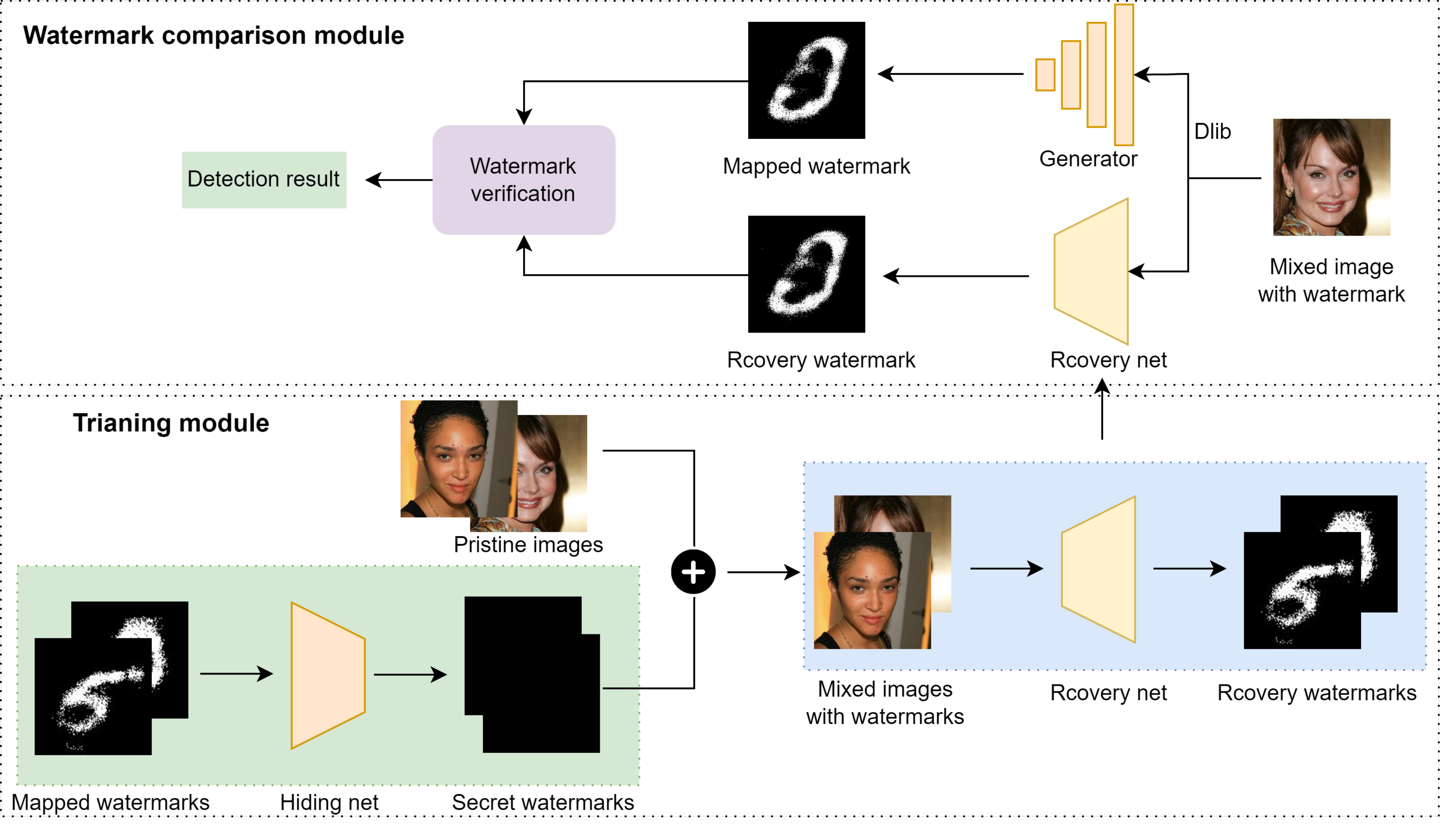}
  \caption{Architecture of Watermark-based verification strategy (WVS).}
  \label{fig:wvs}
\end{figure*}
\subsection{GAN-based One-way Dynamic Watermark Generating Mechanism (GODWGM)}
The GAN-based One-way Dynamic Watermark Generating Mechanism plays a vital role in our proactive deepfake detection framework. This mechanism transforms facial features into associated watermarks through a one-way process, incorporating two essential characteristics. First, its one-way nature prevents attackers from reverse-engineering original image information or reconstructing the generator network, thus maintaining the method's effectiveness. Second, the dynamic nature of the watermarks helps avoid vulnerabilities typically associated with fixed watermarks. It comporises two key components: a \textbf{mapped watermark generating module} and a \textbf{watermark generating training module} (Fig.\ref{fig:godwgm}). We choose to build the watermark generating training module on the basis of WGAN\underline{~}GP\cite{wgangp}. In contrast to DCGAN\cite{dcgan}, WGAN\cite{wgan} or other  GANs, WGAN\underline{~}GP can successfully prevent model collapse by introducing a regular term $GP$ (Gradient Panalty), as shown in the second half of Eq. \ref{eq:wgangp}\cite{wgangp}. This enhancement not only improves training stability but also enables the generation of high-quality images. Consequently, the generated watermarks exhibit sufficient randomness, avoiding the tendency towards similarity that can occur with other methods.
\begin{equation}
  L=\sum_{\tilde{\boldsymbol{x}}\sim\mathbb{P}_g}\left[D(\tilde{\boldsymbol{x}})\right]-\sum_{\boldsymbol{x}\sim\mathbb{P}_r}\left[D(x)\right]+\lambda\sum_{\hat{\boldsymbol{x}}\sim\mathbb{P}_{\hat{\boldsymbol{x}}}}\left[(\|\nabla_{\hat{\boldsymbol{x}}}D(\hat{\boldsymbol{x}})\|_2-1)^2\right]
  \label{eq:wgangp}
\end{equation}
During the training process, the MNIST\cite{mnist} is used as the WGAN\underline{~}GP's dataset. An automated script extracts 60,000 128-dimensional facial feature vectors, which are used as input for the WGAN\underline{~}GP generator. Further details on the training process are provided in Section 4.1. To create dynamic watermarks unidirectionally, the mapped watermark generating module utilizes the trained generator from the watermark generating training module. This approach effectively solves the security vulnerabilities associated with fixed watermarks, as the generator functions as a dynamic watermark generator.
\subsection{Watermark-based Verification Strategy (WVS)}
A crucial component of the proactive detection framework is WVS. By integrating steganography with the watermark generator from GODWGM, we propose an  innovative strategy that supports deepfake detection more practically, which implies that neither party has to worry about any extra work beyond sending and receiving images, like synchronizing watermarks. We design a hiding net composed of U-Net\cite{unet} and SENet\cite{senet}, and a convolutional network as the recovery network. Both the hiding and recovery networks are simultaneously trained in the training unit(Fig.\ref{fig:wvs}) of the WVS. The networks employ 3 followed loss functions to facilitate the training process.

\textbf{Hiding loss}. It utilizes both the pristine image and the watermark image as inputs for the encoder, thereby constraining the similarity between the resulted mixed image and the pristine image through mean squared error (MSE). \begin{equation}\mathcal{L}_{h}=MSE(F_{en}(I,W),I)\label{eq:lh}\end{equation} Where $I$ represents the pristine image, $W$ denotes the watermark, $F_{en}$ signifies the hiding network, and MSE is a function that quantifies the pixel discrepancy between the mixed image and the pristine image.

\textbf{Recovery loss}. By considering the mixed image and watermark as inputs, we evaluate the similarity between the recovered watermark and the original watermark using a MSE constraint. \begin{equation}\mathcal{L}_{r}=MSE\big(F_{de}\big(F_{en}(I,W)\big),W\big)\end{equation} Where $F_{de}$ represents the recovery network.

\textbf{Overall loss}.	$\lambda_1$ and $\lambda_2$ are the weights of the hiding and recovery network loss functions, respectively.
\begin{equation}\mathcal{L}_{loss}=\lambda_1 L_{h}+\lambda_2 L_{r}\end{equation}
In order to retrieve the watermark $W_r$ from the mixed image $I_m$, the recovery network in the training module is first utilized. Next, the mapped watermark of mixed image $W_m$ is generated using generator in GODWGM, and $W_r$ and $W_m$ are sent into the watermark comparison module to compute the cosine similarity.
\begin{equation}\frac{W_{1}\cdot W_{2}}{\|W_{1}\|\times\|W_{2}\|}>\tau \end{equation}
If the watermarks $W_r$ and $W_m$ meet the above formula, the watermarks are judged to be similar, that is, the image is real, otherwise the image is fake or not proactively protected. Where $W_1$, $W_2$ represent the vectors of the watermarks, and $\tau$  is the judgment threshold, which is 0.8.
\section{Experiment}
\label{sec:exper}
\subsection{Implement Details}
\textbf{Preprocess}. To begin with, 60,000 images at 256$\times$256 resolution are selected from the CelebA dataset\cite{celeba} after removing images unsuitable for facial  feature extraction (\eg, occlusion, sideways) using Dlib\cite{dlib}. These 60,000 images are used to train the generator in GODFGM. Furthermore, we designate a subset of 30000 images from this collection for training the hiding and recovery networks in WVS.
\begin{figure}[t]
  \centering
  \begin{subfigure}{0.48\linewidth}
    \centering
    \includegraphics[width=1\linewidth]{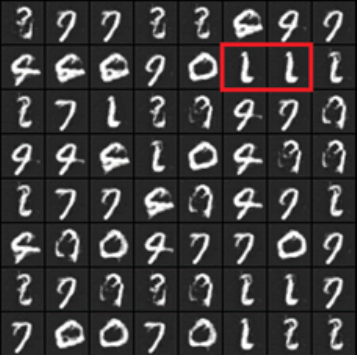}
    \caption{Watermarks generated by DCGAN.}
    \label{fig:dcgan}
  \end{subfigure}
  \hfill
  \begin{subfigure}{0.48\linewidth}
    \centering
    \includegraphics[width=1\linewidth]{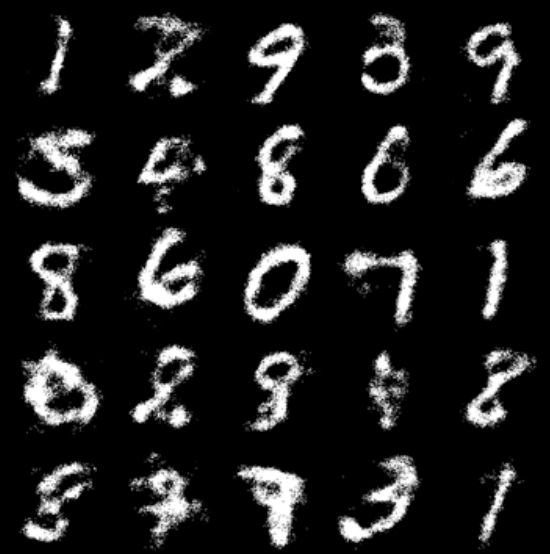}
    \caption{Watermarks generated by WGAN\underline{~}GP.}
    \label{fig:wgangp}
  \end{subfigure}
  \caption{Watermarks generated by WGAN\underline{~}GP and DCGAN. The watermarks in (a) have low similarity, and the watermarks in (b) are similar.}
   \label{fig:gancompare}
\end{figure}

\textbf{Training Details.} (1)\textbf{GAN Training in GODFGM:} Rather than using random noise as the inputs of generator in the WGAN\underline{~}GP\cite{wgangp}, we employ 60,000 128-dimensional feature vectors extracted from the preprocessed dataset of CelebA\cite{celeba} using the Dlib instead. These feature vectors, with each dimension constrained between 0 and 1, are stored in text files using automated scripts and fed into the generator. MNIST is used as the dataset for the discriminator in WGAN\underline{~}GP. The generator outputs are grayscale images of handwriting style. The training epochs and the batch size are set to 100 and 64. The learning rates for the discriminator and generator are set to 0.004 and 0.001. Adam\cite{adam} optimization is used for both networks. An identical procedure is applied to train the DCGAN. Fig.\ref{fig:gancompare} depicts the mapped watermarks obtained from both models. The watermarks generated by WGAN\underline{~}GP are more random and diverse than the ones generated by DCGAN, which makes the watermark verification easier to work and increase the accuracy of detection. 
(2)\textbf{Hiding and Recovery Network Training in WVS.} The hiding network is composed of a U-Net and SENets whose reduction ratio (a hyperparameter of SENet) is 10 according to our experiment results. The recovery network consists of a 6-layer convolutional architecture. Our training dataset consists of 30,000 carrier images and 30,000 MNIST-style facial feature mapped watermarks generated by GODWGM, which serve as the images to be hidden. We set the batch size to 32, learning rate to 0.001, and train for 20 epochs. We employ the Adam optimizer for network training. It is significant to note that this training process does not require image labels.
\begin{figure}[t]
  \centering
  \begin{subfigure}{1\linewidth}
    \centering
    \includegraphics[width=1\linewidth]{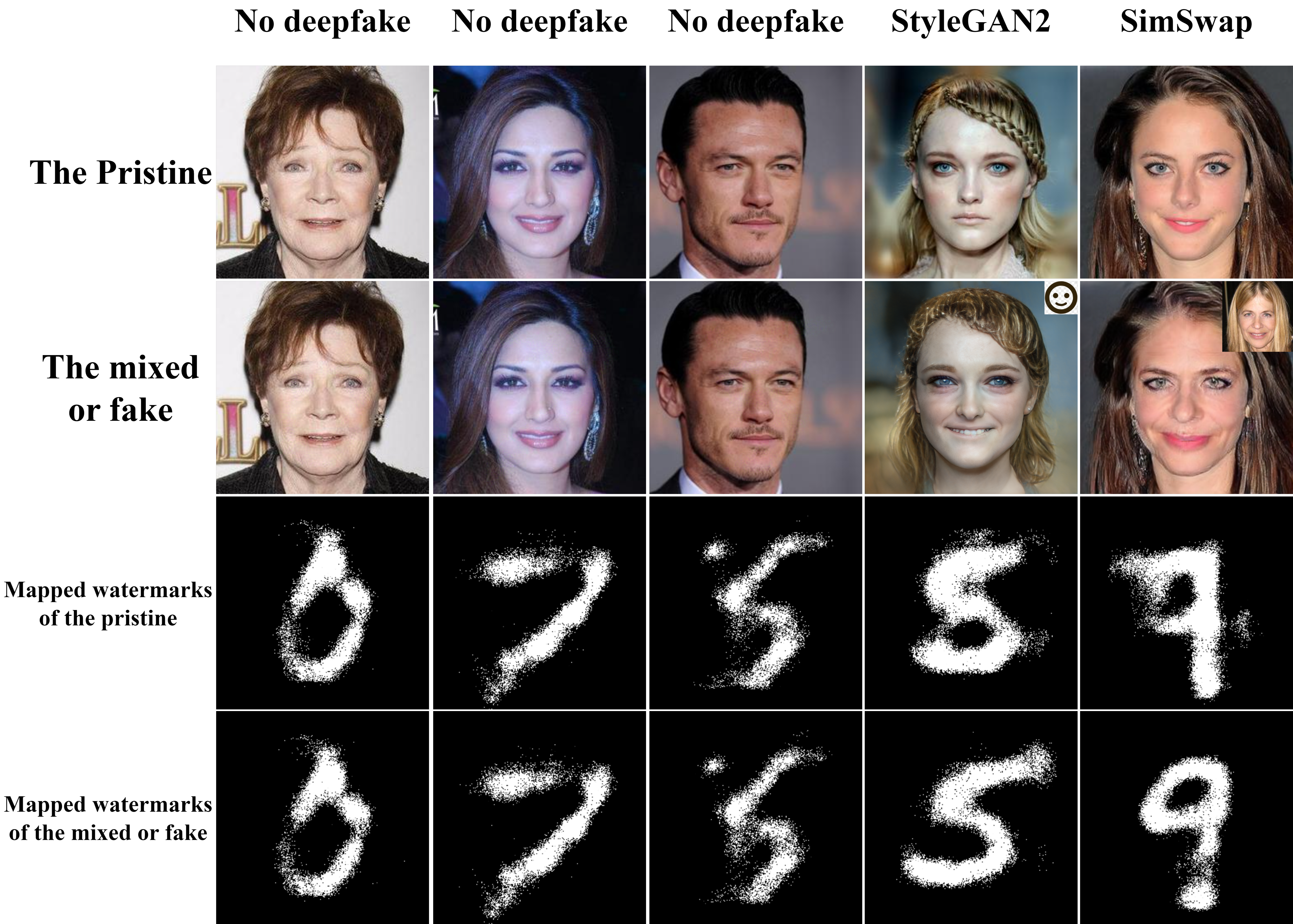}
  \end{subfigure}
  \caption{Images and relevant watermarks.}
  \label{fig:watermark_result}
\end{figure}
\subsection{Experiment Setting}
\textbf{Dataset.} In GODWGM, the training dataset is the MNIST consisting of 600,000 handwritten Arabic numerals with a resolution of 28$\times$28 pixels and an equivalent number of facial feature vectors. In WVS, the dataset to be hidden comprises 30,000 watermarks generated by GODWGM, while the carrier dataset of 30,000 images from CelebA. We employ two identity manipulation methods, InfoSwap\cite{infoswap} and SimSwap\cite{simswap}, as well as two facial attribute editing methods, StyleGAN2\cite{stylegan} and AttGAN\cite{attgan}, to assess the detection performance of our proposed method. For the test dataset, we retain 1000 pristine images of real examples while producing 1000 fake samples for each deepfake method mentioned above.

\textbf{Compared Baselines.} To demonstrate the effectiveness of the proposed method, we evaluate it against several detection methods with publicly available source codes or pre-trained models: SBI \cite{sbi}, CNNS\cite{cnnd} and DDR\cite{bts}. (1)SBI: a novel method that generates synthetic training data to guide the classifier to focus more on the face's global information and enhance the detection model's generalization capacity by using diverse and nearly indiscernible spurious samples. (2)CNNS: an universal classifier capable of identifying fake images generated by CNN networks. (3)DDR: a detection framework that combines low-level and high-level features, achieving more stable detection results when the data distribution changes. These are the latest reproducible passive detection methods. Additionally, we replicate an proactive detection method called RootAttr\cite{agf} to compare with the method proposed in this paper. However, we find that only release the code for steganography based on StegaStamp\cite{sts} is publicly available.

\textbf{Evaluation Metrics.} We employ multiple metrics to assess the similarity between the pristine images and the mixed images, including Peak Signal-to-Noise Ratio (PSNR), Structural Similarity (SSIM), Mean Squared Error (MSE), and Cosine Similarity (CS). These metrics are also used to quantify the similarity among watermarks from various sources. To evaluate the detection performance across different categories of deepfake detection methods, we use Accuracy (ACC), Precision (PREC), F1-Score, and Recall measures. The criterion for judging the true prediction of the method proposed in this paper is that the cosine similarity threshold of the watermarks used for comparison is greater than the $\tau$, 0.8 in this case, based on the experiments we conduct.
\begin{figure}[t] 
  \centering
  \begin{subfigure}{1\linewidth}
    \centering
    \includegraphics[width=1\linewidth]{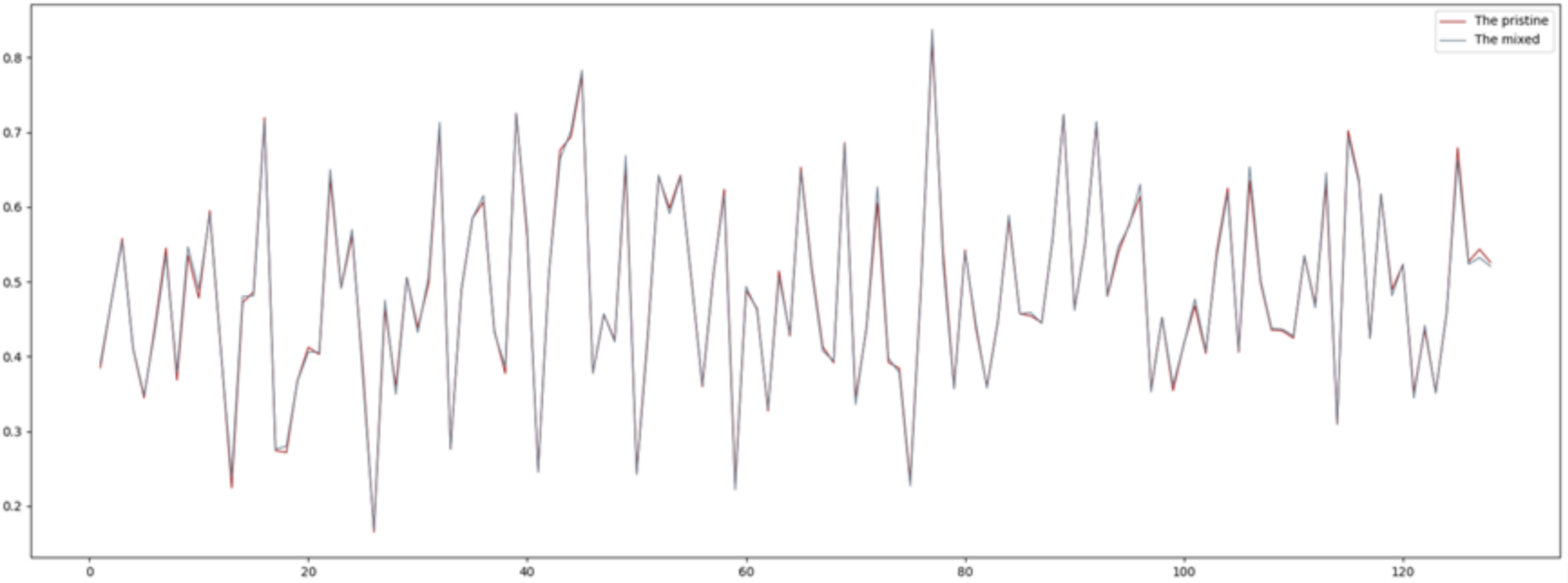}
    \caption{Curves of the pristine and the mixed.}
    \label{fig:fake_curve}
  \end{subfigure}
  \hfill
  \begin{subfigure}{1\linewidth}
    \centering
    \includegraphics[width=1\linewidth]{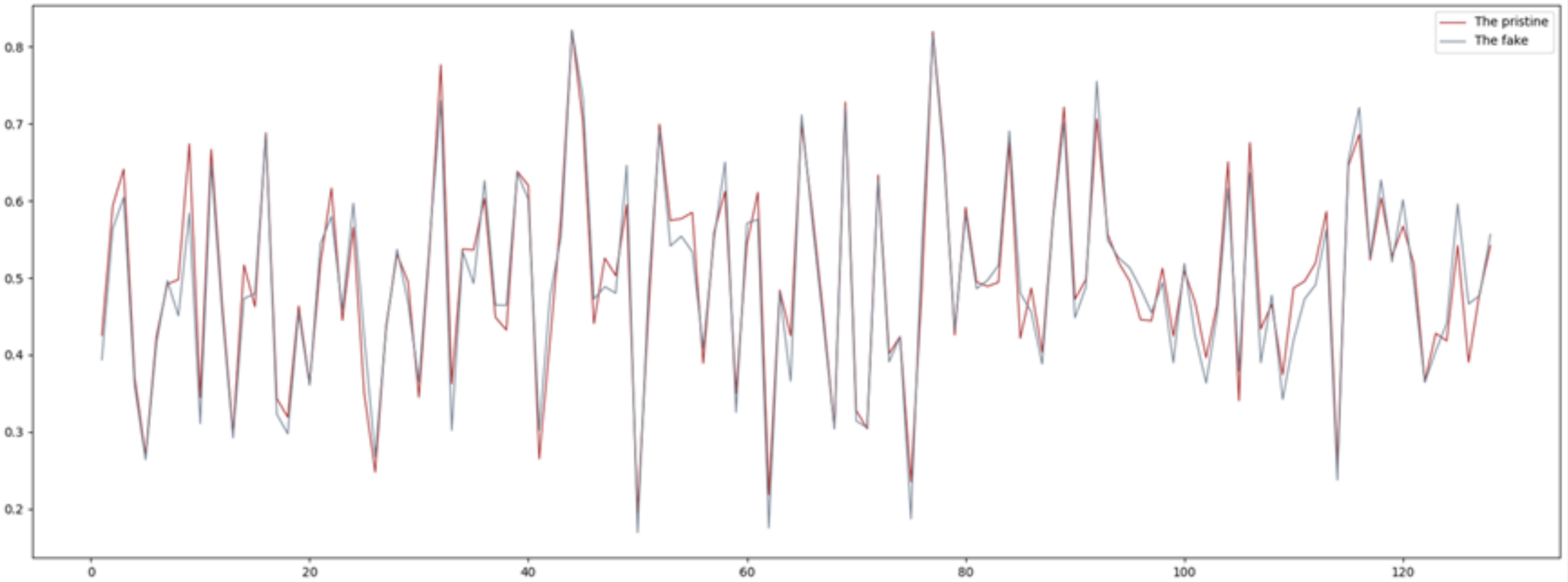}
    \caption{Curves of the pristine and the fake forged by StyleGAN2\cite{stylegan2}.}
    \label{fig:mixed_curve}
  \end{subfigure}
  \caption{Curves of different images. The facial feature curves represented by (b) have greater differences.}
   \label{fig:curve_mixed_fake}
\end{figure}
\subsection{Quantitative Results and Analysis}
\textbf{Feasibility of the Proposed Method.} We randomly select three images(1-3th column in Fig.\ref{fig:watermark_result}) and apply GODWGM to obtain the mapping watermarks of both pristine and mixed images. The visual distinction between pristine images and their watermarked counterparts is imperceptible to the human eye, as is the difference between their associated watermarks. Simultaneously, we apply two deepfake methods to the images in the 4th and 5th columns of Fig.\ref{fig:watermark_result}: SimSwap\cite{simswap} for identity manipulation and StyleGAN2\cite{stylegan2} for facial attribute editing. We then utilize the same procedure to generate relevant watermarks. To demonstrate the effectiveness of our proposed method, we compute the similarity indicators (SSIM, PSNR, MSE, and CS) between the images and watermarks, averaging the results to mitigate randomness. It is evident from the first row of Table \ref{tab:watermark_vs} that the pristine image retains its facial features after the watermark embedding, and the watermarks mapped from both pristine and mixed images remain consistent. The pristine image's and the mixed image's facial feature curves are further drawn in Fig.\ref{fig:curve_mixed_fake}. The facial features are rarely altered in the watermarked image, as can be observed. However, when deepfakes are applied, the results can be seen in row 2 and 3 of Table \ref{tab:watermark_vs}, and the facial feature curves (Fig.\ref{fig:fake_curve}) will also show notable deviations. This is sufficient evidence to support the deepfake detection.
\begin{table}
  \centering
  \begin{tabular}{@{}lcccc@{}}
    \toprule
    Watermark & PSNR$\uparrow$ & SSIM$\uparrow$ & MSE$\downarrow$    & CS$\uparrow$ \\
    \midrule
    P and M   & 42.73 & 0.989 & 3.47    & 0.99996\\
    P and A   & 10.97 & 0.692 & 5197.25 & 0.75647\\
    P and S   & 9.63  & 0.670 & 7070.25 & 0.59223\\
    \bottomrule
  \end{tabular}
  \caption{Similarity metrics between different watermarks. P represents the watermark of the pristine image, while M, A, S represent watermarks of the mixed, fake (StyleGAN2), fake (SimSwap) images respectively.}
  \label{tab:watermark_vs}
\end{table}
\begin{table}
  \centering
  \begin{tabular}{@{}lcccc@{}}
    \toprule
    Watermark       & PSNR$\uparrow$ & SSIM$\uparrow$ & MSE$\downarrow$     & CS$\uparrow$ \\
    \midrule
    Sequence        & 3.40  & 0.15  & 29718.46 & 0.669 \\
    Grayscale Image & 42.33 & 0.96  & 151.26   & 0.989 \\
    \bottomrule
  \end{tabular}
  \caption{Similarity metrics between sequences and grayscale images. Grayscale images outperform sequences in all metrics.}
  \label{tab:img_seq}
\end{table}

\begin{figure}[t]
  \centering
  \begin{subfigure}{0.32\linewidth}
    \centering
    \includegraphics[width=0.98\linewidth]{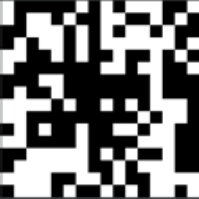}
    \caption{Original sequence}
    \label{fig:origin_seq}
  \end{subfigure}
  \begin{subfigure}{0.32\linewidth}
    \centering
    \includegraphics[width=0.98\linewidth]{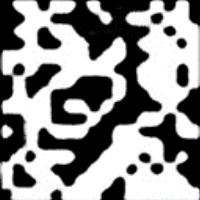}
    \caption{Recovery sequence}
    \label{fig:recovery_seq}
  \end{subfigure}
  \begin{subfigure}{0.32\linewidth}
    \centering
    \includegraphics[width=0.98\linewidth]{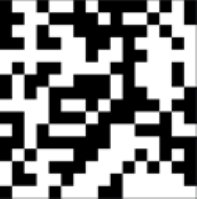}
    \caption{Final sequence}
    \label{fig:final_seq}
  \end{subfigure}
   \caption{Conversion from sequences to grayscales image.}
   \label{fig:seq}
\end{figure}

\textbf{Images are Superior to Sequences.} Instead of employing binary sequences as watermarks, we choose to employ grayscale images because binary sequences are more likely to lose information when embedded directly into images. This choice mitigates potential information loss and enhances the robustness of subsequent watermark recovery. A 256-bit binary sequence is transformed into a 16 x 16 grayscale image, with 0 representing 255 and 1 representing 0 in the grayscale values. The original sequence is shown in Fig.\ref{fig:origin_seq}, whereas the recovered sequence is shown in Fig.\ref{fig:recovery_seq}. The final sequence of the recovered one is produced after transformation. The CS and the MSE values between the original sequence and the final recovered one are 0.669 and 29718.46, respectively, and neither of them can adequately support further detection of deepfake. It is evident from Table \ref{tab:img_seq} that the recovery effect is superior when using grayscale images rather than binary sequences as watermarks.
\begin{table*}[ht]
  \centering
  \begin{tabular}{@{}lcccc@{}}
    \toprule
    \multirow{2}{*}{Method}&\multicolumn{2}{c}{Facial Attribute Editing}&\multicolumn{2}{c}{Identity Manipulation}\\
                     & StyleGAN2      & AttGAN         & SimSwap        & Infoswap\\
    \midrule
    SBI\cite{sbi}    & 0.50/0.50/0.63 & 0.49/0.49/0.61 & 0.50/0.50/0.63 & 0.49/0.49/0.62 \\
    CNNS\cite{cnnd}  & 0.68/0.61/0.75 & 0.75/0.67/0.80 & 0.70/0.62/0.76 & 0.50/0.50/0.67 \\
    DDR\cite{bts}    & 0.88/0.81/0.89 & 0.89/0.83/0.90 & 0.74/0.66/0.79 & 0.49/0.50/0.66 \\ \hline
    Ours             & 0.96/0.93/0.95 & 0.93/0.88/0.91 & 0.99/0.99/0.99 & 0.99/0.99/0.99 \\
    \bottomrule
  \end{tabular}
  \caption{ACC/PREC/F1-Score values in all test datasets we make. Ours has a better result.}
  \label{tab:passive_method_com}
\end{table*}
\begin{table}
  \centering
  \begin{tabular}{@{}lcccc@{}}
    \toprule
    Method & Image size & Watermark Size & PSNR$\uparrow$ & SSIM$\uparrow$ \\
    \midrule
    UDH & 256$\times$256 & 256$\times$256 & 40.38 & 0.981 \\
    Ours & 256$\times$256 & 256$\times$256 & 42.23 & 0.986 \\
    \bottomrule
  \end{tabular}
  \caption{Steganography quality. Watermark is a 256 $\times$ 256 resolution grayscale image  }
  \label{tab:udh_vs}
\end{table}

\textbf{Visual Quality.} Maintaining visual consistency before and after hiding watermarks is an essential requirement of steganography, although it is not the primary focus of this paper. While the goal is not to achieve visual excellence, the proposed method exhibits notable performance compared to other approaches. We achieve the highest SSIM value of 0.986, surpassing the values of current methods such as HiDDeN(0.888)\cite{hidden}{}, MBRS(0.775)\cite{mbrs}, CIN(0.967)\cite{cin}, RootAttr(0.975)\cite{agf}, AntiForgery(0.953)\cite{antiforgery}, CMUA(0.857)\cite{cuma}, and FaceSigns(0.889)\cite{facesign}, while hiding a 256$\times$256 grayscale watermark, which contains significantly more information than a 256-bit binary sequence. In terms of PSNR, we(42.23) outperform HiDDeN(33.26), MBRS(33.01), AntiForgery(35.62), CUMA(38.64), and FaceSigns(36.99), although not as well as IN(43.37), RootAttr(43.93). However, our approach is still better than them to some extent, as our hidden data volume (a 256 x 256 grayscale image) is significantly larger than binary sequences (up to 256 bits). Under the identical settings as indicated in Table \ref{tab:udh_vs}, we duplicated the high-performance steganography algorithm, UDH\cite{udh}. In every metric, the method proposed in this paper perform better than UDH.

\textbf{Performance Comparison.} Since all of the passive detection methods compared in this study claim to enhance the models' capacity for generalization to some degree, we directly utilize the pretrained model made available by the official. As demonstrated in Table \ref{tab:passive_method_com}, the method proposed in this study outperforms the methods tested in terms of both identity manipulation and facial attribute editing across metrics. This implies that the proposed method not only exhibits exceptional detection performance but also shows high generalization ability when addressing different deepfakes.

\section{Limitations}
\label{sec:limit}
The proposed method assumes that watermarked images will retain their watermarks after undergoing common image processing techniques (\eg, compression, Gaussian blur) as well as the majority of deepfakes (\eg, faceswap, facial attribute editing). However, this is not directly addressed in our work. Through the design of loss functions, these processing and forgery techniques can be included during the training phase to enhance the robustness of the watermark. Despite these measures, new deepfake methods will emerge, designed explicitly to circumvent such protections. However, images generated by these new deepfakes typically fail to retain the watermark (which can be viewed as a blank watermark), which suggests that the method proposed in this paper is still effective because the mapped watermark and the blank watermark differ more noticeably.

\section{Conclusion}
\label{sec:conc}
In this paper, we introduce a facial feature-based proactive detection framework. Utilizing the GODWGM framework we developed, a one-way and dynamic mapping of facial features to watermarks is facilitated. The watermark representing the facial features is hidden inside the pristine image by the WVS system we propose. Deepfake detection is then performed by comparing the resemblance of the recovered watermark to that of the mapping watermark within the altered image. Experiment results demonstrate that our method achieves robust performance in detecting deepfakes, particularly in scenarios involving various tampering methodologies. Further research will concentrate on ensuring that watermarks can be successfully recovered after diverse deepfakes or processing.

{
    \small
    \bibliographystyle{ieeenat_fullname}
    \bibliography{main}
}


\end{document}